%% file: main.tex
\documentclass[10pt,twocolumn,letterpaper]{article}

\usepackage{cvpr}              %% To produce the CAMERA-READY version
\usepackage{graphicx}
\usepackage{amsmath}
\usepackage{amssymb}
\usepackage{booktabs}
\usepackage{multicol}
\usepackage{multirow}

\usepackage[pagebackref,breaklinks,colorlinks]{hyperref}

\usepackage[capitalize]{cleveref}
\crefname{section}{Sec.}{Secs.}
\Crefname{section}{Section}{Sections}
\Crefname{table}{Table}{Tables}
\crefname{table}{Tab.}{Tabs.}

\def\cvprPaperID{*****} % *** Enter the CVPR Paper ID here
\def\confName{CVPR}
\def\confYear{2022}

\begin{document}

\title{LLMVA-GEBC: Large Language Model with Video Adapter \\for Generic Event Boundary Captioning}

\author{Yunlong Tang, Jinrui Zhang, Xiangchen Wang, Teng Wang, Feng Zheng\\
SUSTech VIP Lab\\
{\tt\small \{tangyl2019, zhangjr2018, wangxc2019, wangt2020\}@mail.sustech.edu.cn,}
{\tt\small zhengf@sustech.edu.cn}
% \and
% Ping Luo
}

\maketitle

%%%%%%%%% ABSTRACT
\input{sections/abstract}
\input{sections/introduction}
\input{sections/method}
\input{sections/experiments}
\input{sections/conclusion}

%%%%%%%%% BODY TEXT

%%%%%%%%% REFERENCES
{\small
\bibliographystyle{ieee_fullname}
\bibliography{egbib}
}

\end{document}

%% file: sections/abstract.tex
\begin{abstract}
Our winning entry for the CVPR 2023 Generic Event Boundary Captioning (GEBC) competition is detailed in this paper. Unlike conventional video captioning tasks, GEBC demands that the captioning model possess an understanding of immediate changes in status around the designated video boundary, making it a difficult task.
This paper proposes an effective model LLMVA-GEBC (\textbf{L}arge \textbf{L}anguage \textbf{M}odel with \textbf{V}ideo \textbf{A}dapter for \textbf{G}eneric \textbf{E}vent \textbf{B}oundary \textbf{C}aptioning): (1) We utilize a pretrained LLM for generating human-like captions with high quality. (2) To adapt the model to the GEBC task, we take the video Q-former as an adapter and train it with the frozen visual feature extractors and LLM.
Our proposed method achieved a 76.14 score on the test set and won the $1^{st}$ place in the challenge. Our code is available at \url{https://github.com/zjr2000/LLMVA-GEBC}.
\end{abstract}

%% file: sections/introduction.tex
\section{Introduction}
The objective of GEBC is to produce three captions that depict a particular event boundary: one for the subject involved in the event boundary and the other two for the status before and after the boundary~\cite{wang2022geb+}. Recently, Large Language Models (LLMs) have been rapidly developing and becoming increasingly popular and powerful in natural language generation. However, previous methods for GEBC have rarely utilized LLMs for caption generation. In this work, we propose an effective model that utilizes a pretrained LLM and video adapter to generate high-quality captions for the GEBC task.
Our proposed model has three key components: a pretrained LLM, video adapters, and multiple visual feature extractors. We utilize the Open Pre-trained Transformer (OPT)~\cite{zhang2022opt}, a pretrained LLM, to generate captions that are more human-like and accurate. To extract video features, we take the BLIP-2~\cite{li2023blip} model's image encoder (CLIP-ViTG~\cite{zhai2022scaling}) with Q-former as our primary feature extractor. In addition to BLIP-2, we incorporate other features such as CLIP~\cite{radford2021learning}, Omnivore~\cite{girdhar2022omnivore}, and VinVL~\cite{zhang2021vinvl}, to obtain both frame-level and region-level information from the video. The video Q-former~\cite{zhang2023video} is a crucial component of our model as it serves as a video adapter that can convert video features into query tokens that the LLM can understand. By training the video Q-former with the frozen feature extractor and LLM, it can adapt to the GEBC task's data (i.e., videos and corresponding captions) by adjusting its parameters through training. This adaptability makes the Q-former highly flexible and allows the LLM to generate captions specific to the GEBC task without compromising the performance of the LLM.

In summary, our proposed LLMVA-GEBC offers a promising approach to improving the quality of captions generated for the GEBC task by leveraging the power of LLMs and incorporating multiple features to generate more effective captions, which becomes the state of the art on GEBC task. The video Q-former plays a crucial role in our model as it allows the LLM to adapt to the GEBC task and generate captions that are specific to the task without compromising LLM's performance.

%% file: sections/method.tex
\section{LLMVA-GEBC}
\begin{figure*}
\centering
% \vspace{-2em}
  \includegraphics[width=1.0\textwidth]{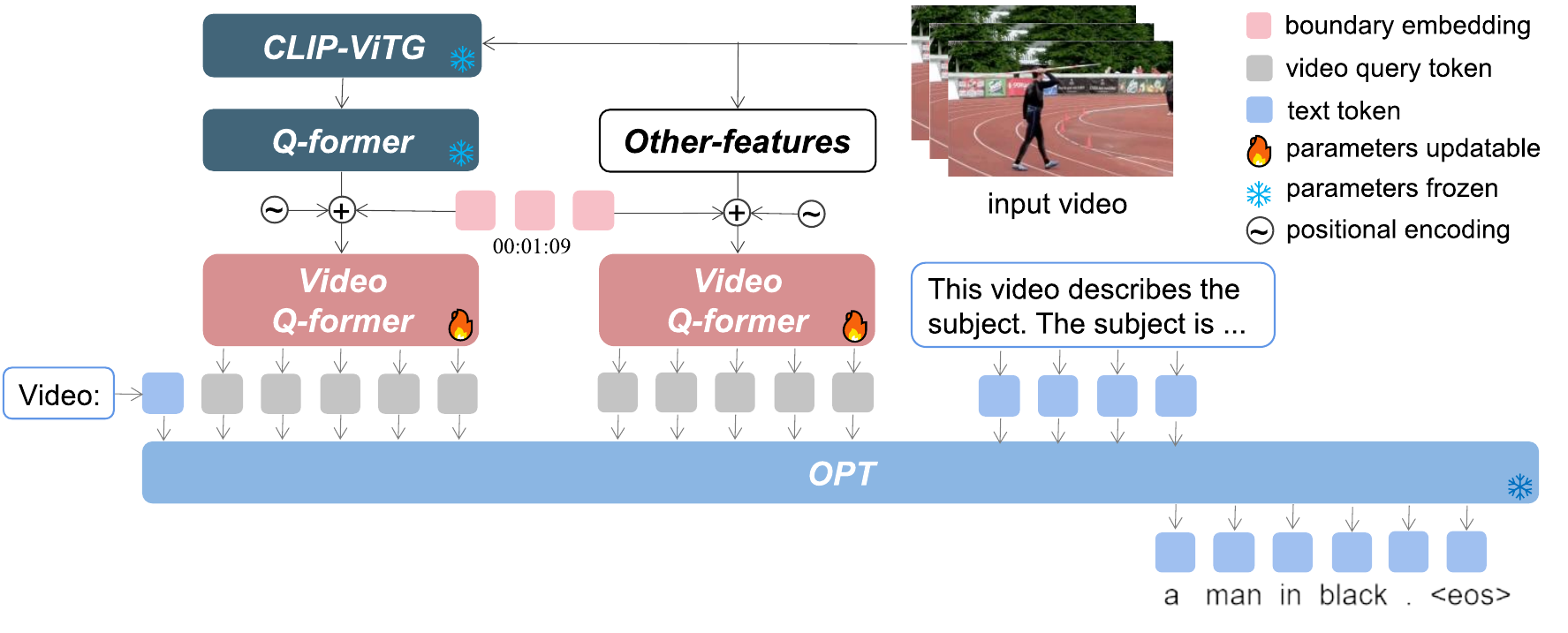}
  \vspace{-2em}
  \caption{Illustration of the proposed LLMVA-GEBC model.}% It shows a process of generating the caption for the Subject in the input video. The captions for Status\_Before and Status\_After can be produced by replacing the word \textit{subject}.
 \vspace{-1em}
  \label{fig:model}
\end{figure*}

GEBC can be formally defined as follows: given a video consisting of $T$ frames and $N$ event boundaries represented by $\{t^n\}_{n=1}^N$, the model should generate three captions for each event boundary: $C_{sub}$, $C_{bef}$, and $C_{aft}$, which describe the subject, the status before the boundary, and the status after the boundary, respectively. Our LLMVA-GEBC model, as illustrated in Figure~\ref{fig:model}, comprises a primary feature extractor CLIP-ViTG with Q-former, various other visual feature extractors, video Q-former adapters, and the LLM, OPT. The parameters of video Q-former adapters are updatable when training, while the feature extractors and LLM are frozen.

\subsection{Feature Encoding}
We first use CLIP-ViTG~\cite{zhai2022scaling} with Q-former to extract BLIP-2~\cite{li2023blip} feature $F_{blip} \in \mathbb{R}^{L_0\times q_0 \times d_0}$ from the input video as the primary visual feature, where $L_0$ is the number of frames sampled from all $T$ frames with a stride of $m$, $q_0$ is the number of the query tokens from Q-former for each frame, and $d_0$ is the dimension of the primary visual feature.

We extract other visual features as supplements from video using multiple pretrained models, including CLIP \cite{radford2021learning} and Omnivore \cite{girdhar2022omnivore}. We extract the two frame-level features by sampling a sequence of frames with a stride of $m$ and then extract features for each sampled frame. To enable batch processing, we resize the temporal dimension of the features to a fixed number $L$ using an interpolation or downsampling strategy. The processed primary visual feature will be $F_{blip} \in \mathbb{R}^{L\times q_0 \times d_0}$. The processed frame-level features are denoted by $F_{clip} \in \mathbb{R}^{L \times 1 \times d_1}$ or $F_{omni} \in \mathbb{R}^{L \times 1 \times d_2}$, where $d_1$ and $d_2$ are the hidden dimensions of the feature extractors. We also extract region-level features using VinVL \cite{zhang2021vinvl}. To align the dimension, the region-level features are denoted by $F_{vinvl} \in \mathbb{R}^{L \times N_o \times d_3}$, where $d_3$ is the hidden dimension of the feature extractors, and $N_o$ represents the number of objects with high confidence in each frame. All the dimensions of other visual features will be projected to $d_0$.

\subsection{Video Query Token Generation}
The extracted visual features will be used to generate video query tokens. Specifically, learnable positional encoding will be added with both primary and other visual features. To establish a connection between visual features and event boundaries, we also generate boundary embedding vectors for each boundary. The time boxes of the event boundaries $\{t^n\}_{n=1}^N$ are normalized to the range $[0, 1]$ and then passed through an inverse Sigmoid function. Next, we apply positional encoding and layer normalization to convert them into vectors. Finally, all the time boxes are projected onto $\mathbb{R}^{d_0}$ to get the boundary embedding vectors, which will be added with the visual features.

Then we feed the position-encoded and boundary-encoded primary visual features to video Q-former~\cite{zhang2023video}, whose architecture is the same as Q-former in BLIP-2~\cite{li2023blip}, get the video embedding vectors, and incorporate a linear layer to transform the video embedding vectors into video query tokens $V_0\in\mathbb{R}^{q_0\times h}$ for compatibility with the input of the following LLM, where $h$ is the dimension. A similar operation is applied to position-encoded and boundary-encoded other visual features, and corresponding video query token $V_1\in\mathbb{R}^{q_1\times h}$ will be obtained, where $q_1$ is the number of video query tokens from other visual features. 

\subsection{Caption Prediction}%``Video:''
We utilize an LLM to generate captions by feeding a soft prompt that is constructed with video query tokens and text tokens. Specifically, we first concatenate the token of a text prefix $P$ with the video query tokens $V_0, V_1$, and a suffix $S(A)$ to obtain the prompt:
\begin{equation*}
\begin{alignedat}{2}
&Prompt = (\text{``Video:''},~V,~S(A,B)),\\
&S(A,B) = \text{``This video describes the}~\{A\}.~\{B\}~\text{is''},
\end{alignedat}
\end{equation*}
where $A\in\{\text{``subject''}, \text{``status before''}, \text{``status after''}\}$ and $B\in\{\text{``The subject''},~\text{``Status before change''},~\text{``Status after}$ $\text{change''}\}$.
To perform auto-regressive sentence generation, we use OPT~\cite{zhang2022opt} as our caption generator. The constructed prompts will be the input of the LLM:
\begin{equation*}
    C_{i}=LLM(Prompt), \text{where}~i\in\{sub, bef, aft\}. 
\end{equation*}
The frozen LLM generates one word at a time until it reaches a maximum length of $M$ or generates a special token $<$end$>$. During training, we use the cross-entropy between the predicted word probability and the ground truth to train the model.

% \begin{equation*}
% \begin{alignedat}{2}
% &Prompt = (\text{``Video:''},~V,~S(A,B)),\\
% &S(A,B) = \text{``This video describes the}~\{A\}.~\{B\}~\text{is''},
% \end{alignedat}
% \end{equation*}
% where $A\in\{\text{``subject''},~\text{``status before''},~\text{``status after''}\}$ and $B\in\{\text{``The subject''},~\text{``Status before change''},~\text{``Status after change''}\}$.
% \begin{equation*}
%     Completion_{i}=LLM(Prompt), \text{where}~i\in\{sub, bef, aft\}. 
% \end{equation*}

%% file: sections/experiments.tex
\section{Experiments}

\subsection{Implementation Details}
We utilized the pre-trained weights of BLIP-2 for CLIP-ViTG with Q-former, and the weights pre-trained on webvid2M by Video-LLaMA~\cite{zhang2023video} for video Q-former. For LLM, we used the weights of pretrained OPT-13B with 13 billion parameters. We use the full training set to train the model. The object number per frame $N_o$ is set to 50. Zero padding is added if the object number is less than $N_o$. We take top $N_{o}$ objects sorted by their confidence if one frame has more than $N_o$ objects. The frame sample interval $m$ is set to 12, 8, and 16 for BLIP-2, CLIP, and Omnivore, respectively. The maximum sentence length $M$ is set to $96$ for all three kinds of captions. We train and validate our method on the Kinetic-GEBC dataset~\cite{wang2022geb+}. The optimizer is AdamW~\cite{loshchilov2017decoupled}. The weight decay and mini-batch size are set to $0.001$ and $16$, respectively. The initial, minimum, and warm-up learning rates are set to $8 \times e^{-5}$, $1 \times e^{-5}$, and $1 \times e^{-6}$, respectively. The max number of epochs is 5.

\subsection{Quantitative Results}
From Table~\ref{tab:eval}, we can see that our LLMVA-GEBC outperformed the baseline ActBERT-Revised~\cite{wang2022geb+} on all evaluation metrics with a relative improvement of 86.6\% ($\frac{76.14 - 40.80}{40.80}$). It also achieved higher scores on all of the average, SPICE, ROUGE\_L, and CIDEr metrics, outperforming the champion solution~\cite{gu2022dual} and the second place~\cite{zhang2022exploiting} for CVPR2022 GEBC competition. This improvement can be attributed to our model's ability to leverage video adapters to unlock the capacity of LLM to generate better captions in terms of semantic similarity, linguistic quality, and diversity. Overall, our results demonstrate the effectiveness of our LLMVA-GEBC model for the GEBC task. We also try different combinations of other features as shown in Table~\ref{tab:other}, which shows when using CLIP, Omnivore, and VinVL as the other features, we will get the best performance.
\begin{table}[]
\centering
\caption{The final results on the Kinetic-GEBC test set.}
\label{tab:eval}
\begin{tabular}{l|cccc}
\hline
Model                      & AVG            & SPICE          & ROUGE\_L       & CIDEr           \\ \hline
\begin{tabular}[c]{@{}l@{}}ActBERT-\\revised~\cite{wang2022geb+}\end{tabular}                  & 40.80          & 19.52          & 28.15          & 74.71           \\ \hline
\begin{tabular}[c]{@{}l@{}}Context-\\GEBC~\cite{zhang2022exploiting}\end{tabular}   & 72.84          & -          &  -        & -         \\ \hline
\begin{tabular}[c]{@{}l@{}}Dual-Stream\\ Xfmr~\cite{gu2022dual}\end{tabular}   & 74.39          & 33.86          & 41.19          & 148.13          \\ \hline
\begin{tabular}[c]{@{}l@{}}LLMVA-\\GEBC (Ours)\end{tabular} & \textbf{76.14} & \textbf{34.76} & \textbf{41.93} & \textbf{151.72} \\ \hline
\end{tabular}
\end{table}

\begin{table}[]
\centering
\caption{Experiments different combinations of other features on the Kinetic-GEBC validation set.}
\label{tab:other}
\begin{tabular}{l|cccc}
\hline
other feature(s)                                                                  & AVG   & SPICE & ROUGE\_L & CIDEr  \\ \hline
\begin{tabular}[c]{@{}l@{}}Without any\\ other feature\end{tabular}               & 73.24 & 33.63 & 41.26    & 144.83 \\ \hline
InternVideo~\cite{wang2022internvideo}                                                                       & 74.67 & 34.06 & 41.80    & 148.15 \\ \hline
\begin{tabular}[c]{@{}l@{}}InternVideo\\ + CLIP~\cite{radford2021learning}\\ + Omnivore~\cite{girdhar2022omnivore}\\ + DinoV2~\cite{oquab2023dinov2}\end{tabular} & 75.76 & 34.41 & 42.13    & 150.75 \\ \hline
\begin{tabular}[c]{@{}l@{}}InternVideo\\ + CLIP\\ + Omnivore\end{tabular}           & 75.79 & 34.45 & \textbf{42.41}    & 150.52 \\ \hline
\begin{tabular}[c]{@{}l@{}}CLIP\\ + Omnivore\\ + VinVL~\cite{zhang2021vinvl}\end{tabular}                 & \textbf{76.62} & \textbf{34.65} & 42.28    & \textbf{152.94} \\ \hline
\end{tabular}
\end{table}

\subsection{Qualitative Results}
\begin{figure}[h]
\centering
% \vspace{-2em}
  \includegraphics[width=0.5\textwidth]{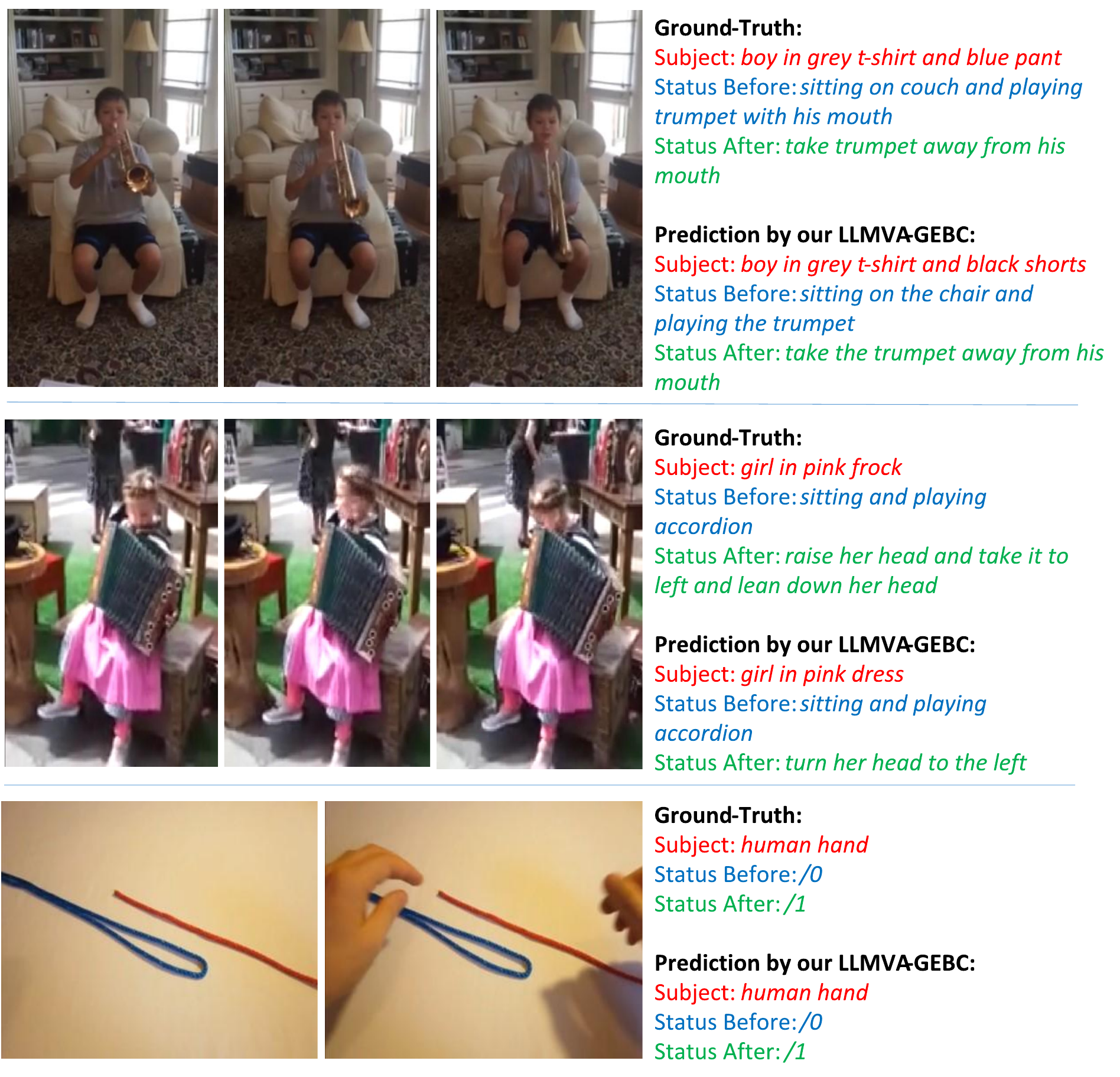}
  % \vspace{-2em}
  \caption{Qualitative Results on Kinetic-GEBC validation set. Our model produces captions of high quality.}
%  \vspace{-1em}
  \label{fig:result}
\end{figure}

Figure~\ref{fig:result} displays qualitative results where our model generates three types of descriptions for each video boundary: Subject, Status Before, and Status After. The captions generated by our model are of high quality.

%% file: sections/conclusion.tex
\section{Conclusion}
Our proposed LLMVA-GEBC model unlocks the ability of LLM on generic event boundary captioning with the video adapters, which achieves a 76.14 score on the Kinetic-GEBC test set, outperforming the official baseline method and the winner method of the CVPR2022 Generic Event Boundary Captioning competition.